# Evaluation of a grammar of French determiners

Éric Laporte

Université Paris-Est – Laboratoire d'Informatique de l'Institut Gaspard-Monge
5, bd Descartes – 77454 Marne-la-Vallée CEDEX 2 – France
eric.laporte@univ-mlv.fr

*Abstract.* *Existing syntactic grammars of natural languages, even with a far from complete coverage, are complex objects. Assessments of the quality of parts of such grammars are useful for the validation of their construction. We evaluated the quality of a grammar of French determiners that takes the form of a recursive transition network. The result of the application of this local grammar gives deeper syntactic information than chunking or information available in treebanks. We performed the evaluation by comparison with a corpus independently annotated with information on determiners. We obtained 86% precision and 92% recall on text not tagged for parts of speech.*

## 1. Introduction[1]

The coverage of existing syntactic-semantic grammars of natural languages is far from complete, but even so, such grammars are complex objects and their construction takes many years. Therefore, it is desirable to assess the quality of parts of a grammar and to control their evolution before it is complete. To date, partial grammars that have been submitted to evaluation are mainly grammars for named entity (NE) recognition or for chunkers. Such evaluation is motivated, in the case of NE recognition, by the existence of direct applications, namely information retrieval (IR) and information extraction (IE); in the case of chunking, it is motivated by the application to syntactic annotation of corpora; in both cases, by the application to shallow parsing. Partial grammars are not confined to IR, IE or shallow parsing. Recent projects have produced parts of deep syntactic grammars, devoted to e.g. determiners. Motivations for constructing deep syntactic grammars involve applications of syntactic parsing, e.g. translation, and also the construction of treebanks.

In this paper, we report an evaluation of a partial syntactic-semantic grammar of French: a grammar of determiners, including complex determiners and combinations of determiners. This grammar neglects dependencies between the determiner and the noun. It takes the form of a recursive transition network (RTN). As compared to chunking, the syntactic information obtained by the application of the grammar is deeper, since the grammar describes complex determiners which may contain several chunks. The output of the parser was compared to a corpus independently annotated with information on determiners. This article is organised as follows. The next section surveys related work. In section 3, we describe the grammar of determiners. Section 4 reports how the grammar was evaluated. We present and analyse the results in section 5.

---

[1] This work has been supported by CNRS and by Senior Planet Co.

## 2. Related work

In recent campaigns of evaluation of syntactic grammars [Briscoe et al. 2002], [Gendner et al. 2003], [Paroubek et al. 2006], each grammar was assessed globally. Evaluation consisted in comparing the output of the parser to a treebank, and no evaluation of separate parts of grammars was organised. However, parts of a manually constructed grammar have not necessarily the same author or the same quality, and are not necessarily built at the same time. Therefore, it is also desirable to assess the quality of parts of a grammar and to control their evolution during their construction.

Partial grammars that have been submitted to evaluation are mainly grammars for NE recognition [Humphreys et al. 1998], [Maynard et al. 2001], [Bick 2004], [Piskorski 2004], for chunkers [Abney 1996], or for both [Saetre 2004][2]. The grammatical formalisms used for these tasks are usually regular expressions (RE) [Piskorski 2004], transducers manually contructed in the form of RE-like formulae [Abney 1996], specific formalisms designed for particular linguistic phenomena [Das et al. 2005], TAGs [Hockey and Mateyak 2000] or RTNs [Nam and Choi 1997], [Senellart et al. 2001], [Saetre 2004]. The symbols recognising words in these grammars are either lexical words or variables which are equivalent to feature structures and refer to various features provided by lexical analysis. Evaluation is performed by comparing the output of the parser to a corpus which has been independently annotated for NEs or chunked.

Partial grammars expressed in the form of RTNs are usually called 'local grammars' [Gross 1997]. The RTN formalism is adapted to NE recognition [Nam and Choi 1997] and chunking [Poibeau 2006] but also to deep syntactic parsing or annotation [Venkova 2000], [Danlos 2005], [Fairon et al. 2005], [Blanc and Constant 2005]. In the recent years, several projects have been devoted to the design and construction of local grammars as components of deep syntactic grammars. Examples of such local grammars deal with: determiners in French, including complex determiners and combinations of determiners [Gross 2001], [Silberztein 2003]; sequences of verbs in French [Gross 1998-99] and Portuguese [Ranchhod et al. 2004]; coordinated noun phrases in Serbo-Croatian [Nenadic 2000]; a general-purpose set of local grammars for constituents such as noun phrases and other clause elements in English, used to recognise syntactic constructions of verbs [Mason 2004]; noun phrases with predicative head in French [Laporte et al. 2006].

Such projects are instances of a bottom-up approach to the construction of deep syntactic grammars. Their objective is (i) to represent the respective syntactic constructs with maximal recall, and (ii) to resolve syntactic ambiguity, but only when this is possible without exploring the context of these constructs[3]. In a syntactic grammar, the resolution of syntactic and part-of-speech (POS) ambiguity is ultimately obtained by the combination of all components, and is not the problem addressed by a single component. Thus, precision is less relevant than recall in the assessment of a component of a syntactic grammar. Quantitative evaluation of such local grammars is particularly useful for the validation of this approach. It is also an indication about the usability of these resources in other projects. However, only three of the contributions listed above report corpus-based evaluation. [Danlos 2005] claims 97% accuracy on a corpus of about 240,000 words. [Silberztein 2003] reports 100% recall on a sample of about 4200 words, but as regards precision, mentions only that it is 'very low'. [Gross 1998-1999] claims 99.8% precision, but does not give the size of the evaluation corpus, nor an assessment of recall. Therefore, very few quantitative data about the coverage of local grammars are presently available. We provide such data referring to a grammar of French determiners.

---

[2] An alternative to the use of grammars for the same tasks is the training of a probabilistic model on an annotated corpus, as has been done for shallow parsing [Sha and Pereira 2003] and named entity recognition [Li and McCallum 2003]. However, these techniques are less compatible with the introduction of syntactic-semantic information and with the recognition of recursive structures.

[3] Recall that RTNs are equivalent to *context-free* grammars.

## 3. The grammar

The grammar is a description of French determiners, including complex determiners and combinations of determiners. We developed it manually from three existing RTNs. It is freely available on the GraalWeb library[4]. In this section, we delimit the scope of the grammar and report how it was constructed.

### 3.1. Scope

In language engineering and traditional grammar, determiners are usually viewed as a POS rather than as a syntactic notion. This view is only a simplification. Determiners behave according to a complex syntax[5]. Some of them are employed with prepositions, e.g. in *beaucoup de facteurs* 'plenty of factors'. Some combine together, as in *les sept pays* 'the seven countries'. In French, the interaction between the frequent preposition *de* 'of' and determiners involves complex rules: for example, the combination of *de* with the plural indefinite article *des*, e.g. in *des mesures* 'measures', produces *de* [Gross 1967], as in the surface form *sous l'effet de mesures draconiennes* 'under the effect of draconian measures' which is observed instead of the expected sequence *sous l'effet de des mesures draconiennes*. Some noun phrases behave as determiners of other nouns, as in *une partie des prêts* 'part of the loans'. Since most of such noun phrases comprise a determiner in turn, sequences that behave as determiners are embedded in others. We do consider such sequences as (generalised) determiners. We refer to nouns such as *partie* 'part' by the term 'determinative nouns'. The scope of the grammar is to describe generalised determiners, defined by [Silberztein 2003] as follows: if each noun phrase is assigned a head noun on syntactic and semantic grounds, the (generalised) determiner of the noun is the sequence from the beginning of the noun phrase to the head noun, excluding the head noun itself and possible adjectives directly attached to the head noun. Thus, in *restituer une partie des prêts* 'give back part of the loans', selectional restrictions point to *prêts* 'loans', rather than to *partie* 'part', as the object of *restituer* 'give back'; therefore, the determiner of the noun phrase *une partie des prêts* is the sequence *une partie des* 'part of'. The scope of our grammar also includes the prepositions *à* and *de* when they introduce the noun group. Thus, the grammar describes the interaction between these prepositions and determiners: contractions (e.g. the surface form *au* standing for *à le* 'to the', or *de* for *de des*) and elisions (e.g. *d'un* for *de un* 'of a'). The sequences described in the grammar are surface forms such as *au*, and not normalized forms such as *à le*. Predeterminers are considered as parts of the corresponding determiners, as *même* 'even' in *même les grandes avenues* 'even the large avenues', except if they are separated from the determiner by a preposition, as in *même dans les grandes avenues* 'even in the large avenues'. The grammar describes syntactic and lexical constraints between elements of (generalised) determiners, e.g. *plusieurs* 'several' is compatible with *la moitié de* 'a half of' but not with *chacun de* 'each of'.

---

[4] http://igm.univ-mlv.fr/~mconstan/library/index_graalweb.html, [Constant 2004].

[5] Most linguistic analyses underlying our work are borrowed from [Gross 1977].

However, the grammar does not specify morpho-syntactic agreement in gender and number, either between the determiner and the noun, or between the determiner and other elements of the sentence (e.g. the subject-verb agreement). This exclusion is motivated by the fact that the parser that we used, the Unitex parser [Paumier 2006], does not support unification in its present version[6]. Determiners occurring without a head noun are also outside the scope of the grammar. For instance, *plusieurs* 'several' can be a syntactic variant of *plusieurs objets* 'several objects'. In that case, the deletion of the head noun is not accompanied by formal modifications of the determiner, but it is in other cases, e.g. in *beaucoup* 'many' for *beaucoup d'objets* 'many objects'.

**3.2. Method of construction of the grammar**

The grammar has been developed manually from three existing RTNs: two grammars of French determiners [Gross 2001], [Silberztein 2003] and a grammar of numerical expressions [Constant 2000]. We removed from Silberztein's grammar two elements: (i) gender and number agreement[7]; (ii) the constraints involving the countable vs. uncountable feature of nouns, since this feature is absent from available lexicons of French. We introduced into the grammar various elements of Gross' and Constant's grammars[8]. From Gross' grammar, we extracted lists of modifying adverbs, of negative adverbial determiners (e.g. *jamais de* 'never any'), of adjectives that can modify determinative nouns, and of adjectives with properties of determiners (e.g. *premier* 'first'). From Constant's grammar, we extracted the description of physical magnitudes and of approximate numerical expressions. Then we enhanced the grammar with more constructions and more constraints, using the same two approaches as Gross, Silberztein and Constant to construct their grammars: the corpus-based bootstrapping method [Gross 2000] and introspection. For example, we introduced combinations of adverbial determiners such as *un peu de* with adjectival determiners such as *chaque*. We also described constraints between successive determinative nouns, as in *trois sortes de parties de* 'three kinds of parts of'[9].

---

[6] We plan to introduce agreement constraints for a new version that we will use with the unification-compatible Outilex parser [Blanc and Constant 2006].

[7] In Silberztein's grammar, agreement is represented by the existence of 4 versions of the grammar for the 4 combinations of the two genders and the two numbers; this redundancy makes the grammar difficult to maintain.

[8] We thank Anastasia Yannacopoulou for her valuable contribution to this work.

[9] We mentioned above that the sequences described in the grammar are surface forms such as *au*, and not normalized forms such as *à le*. However, during the construction of the grammar, we managed all the graphs in the normalized form, and we changed them to the surface form at the end of the construction,

### 3.3. Structure

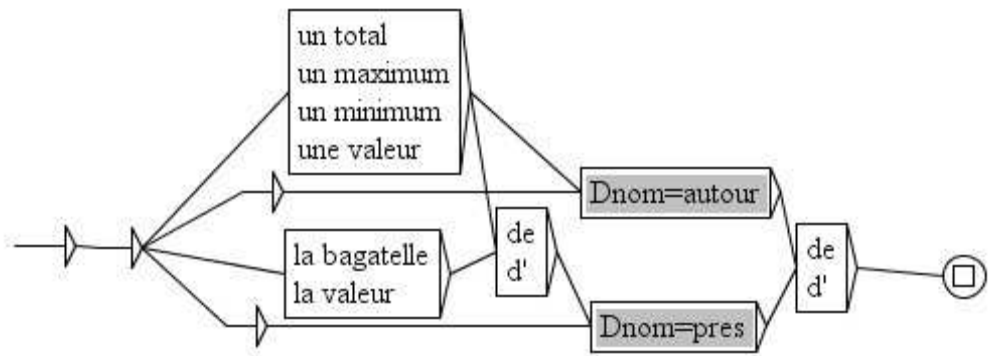

**Figure 1. Graph 'Dnom=presDe' from the local grammar**

The grammar is a network of 186 graphs. One of them is displayed in Fig. 1. There are 3 main graphs: *aDet* and *deDet* for determiners preceded respectively by the prepositions *à* and *de*; *Det* for determiners not preceded by prepositions or preceded by other prepositions. The compilation of these main graphs produces automata with respectively 2143, 2223 and 2044 states. The grammar is strongly lexicalised: it contains 1206 lexical tokens. The grammar recognizes embedded constructs, for instance sequences with several determinative nouns (cf. 3.1). All recursion could be represented in a finite-state way. However, if it is done automatically, through the options of the Unitex grammar compiler, parsing with the resulting grammar is slower; and we checked that if it were done manually, the resulting grammar would be less readable.

### 4. Method of evaluation

Syntactic annotation of text is usually evaluated by comparison with reference treebanks. However, annotation derived from manually constructed grammars is often richer than the information found in golden standards. For example, the annotation guidelines of the Penn Treebank [Marcus et al. 1993] analyse *a boatload of samurai warriors* in the same way as *a conflict between samurai warriors*, e.g. without any information about the quantitative function of *a boatload of*[10]. With this analysis, the head of the noun phrase is *boatload*, in contrast with our phrase structure in which *a boatload of* is a determiner, and *warriors* is the head (cf. 3.1). Our phrase structure is equivalent to additional semantic information, since it is more consistent with that of the majority of noun phrases, where the head is also a semantic head, and the determiner denotes quantitative or identificational information. One of the motivations for building local grammars is the perspective of using them for the construction of treebanks with more informative syntactic-semantic annotation. Therefore, evaluation of our grammar by comparison to standard treebanks would have been inappropriate or even misleading.

---

because this operation obfuscates considerably the grammar and makes it difficult to maintain. We saved the normalized version so that maintenance operations can be performed on it.

[10] Similarly, the French Treebank [Abeillé and Barrier 2004] analyses *J'ai appris un certain nombre d'exigences administratives* 'I got aware of a certain number of administrative requirements' with *nombre* 'number' as the head noun of the complement the verb.

In order to assess the quality of the grammar, we annotated a corpus with information on determiners, we ran the parser with the grammar on the raw version of the evaluation corpus, and we compared the output of a parser with the manual annotation. The evaluation corpus is made of journalistic texts from the newspaper *Le Monde* (1994). Its size is 8000 words. It will be made freely available on the web when this work is published.

### 4.1. Annotation guidelines

The evaluation corpus was annotated with XML tags in order to delimit the (generalised) determiners as defined in 3.1 above. The annotators were given the following guidelines.

Prepositions *à* 'to' and *de* 'of' immediately preceding a determiner are included in the delimited sequence. Other prepositions are not included. The XML tag is respectively <ad> or <dd> instead of <d> if the preposition was included. In case of a compound preposition ending in *de*, only the ending *de* is included in the sequence. For instance, *vis-à-vis de l'Est* 'towards the East' is annotated *vis-à-vis <dd>de l'</dd>Est*. When no determiner occurs between the preposition and the head noun, as in *un changement de concept* 'a change of concept', no annotation is inserted, except in two cases:

- if *de* is analysed as an indefinite determiner, as in *obtenir <d>de</d> meilleures conditions* 'obtain better conditions';

- if *de* is analysed as the surface form corresponding to an underlying sequence of the preposition *de* and an indefinite determiner, as in *sous <d>l'</d>effet <dd>de</dd> mesures draconiennes* 'under the effect of draconian measures'.

Numbers written in figures are annotated in the same conditions as numbers written in letters: <d>100 000</d>, <d>dix</d>.

Determiners occurring without a head noun are not annotated: compare *<d>peu de</d> temps* 'little time' with *Beaucoup semblent d'ailleurs commencer à la comprendre* 'Many indeed seem to begin to understand it'. Percentages not explicitly followed by a head noun are not annotated either: compare *qui couvre <d>environ 8 % des</d> besoins* 'that covers about 8% of the needs' with *accroitre de 40 % ses exportations* 'increase its exports by 40%'.

Determiners inside multi-word units are annotated only if they obey the general syntax of determiners. For example, the determiner is annotated in *<dd>de l'</dd>ordre de 9 à 10 %* 'about 9 to 10%', but not in *rectification d'ordre sémantique* 'correction of a semantic nature'.

The application of these guidelines led to the annotation of 1512 occurrences of determiners: 63% with <d>, 27% with <dd> and 11% with <ad>.

### 4.2. Parsing experiment

We ran a test transducer invoking the 3 main graphs *Det, aDet* and *deDet* on the raw, untagged version of the evaluation corpus, with the Unitex system (version 1.2). We wrote and used this test transducer, instead of using directly the 3 main graphs, in order to mitigate the influence of lexical ambiguity on the results[11]. The test transducer produces an output text which consists of the input text with XML tags inserted before and after each determiner recognised by the local grammar. The XML tags identify whether the sequence was recognised by graph *Det, aDet* or *deDet*[12]. The parsing of the evaluation corpus on a Windows-XP PC took 12s, among which 10s were dedicated to the compilation of the grammar. With a Windows-2000 PC, Unitex parsed 20,452,000 words in 168 mn, which corresponds to 2029 words/second.

### 4.3. Comparison protocol

The annotation inserted by the parser was compared to the manual annotation. The annotation of a sequence in the two files was considered to agree only if both the opening tag and the closing tag occurred at the same place. Two comparisons were performed. In the first one, the three kinds of tags *<d>, <ad>* and *<dd>* were confused: for example, an annotation with *<d>* in the output of the parser was considered to agree with an annotation of the same sequence with *<dd>* in the reference corpus. In the second comparison, two annotations were considered to agree only if the value of the tag was the same.

### 5. Results

We computed the precision (proportion of sequences annotated in the reference corpus among those annotated by the parser) and the recall (proportion of sequences annotated by the parser among those annotated in the reference corpus). The results of the comparison are displayed in Table 1. The 'All' column corresponds to the comparison in which the three kinds of tags are considered equal.

**Table 1. Comparison between parser output and manual annotation**

|           | All | *Det* | *aDet* | *deDet* |
|-----------|-----|-------|--------|---------|
| Precision | 86% | 72%   | 97%    | 35%     |
| Recall    | 92% | 93%   | 91%    | 20%     |

---

[11] Since the evaluation corpus is not tagged for parts of speech, the parser matches variables with text words on the basis of their features found in the lexicons of the system. We used the Dela lexicons [Courtois 1990]. The large coverage of these lexicons tends to lower the precision in the recognition of words and constructions. However, this effect is mechanically mitigated by the length of the sequences described in local grammars: the longer the sequences, the smaller the influence of lexical ambiguity on precision. Since determiners employed without a noun were outside the scope of the experiment, we wrote a test transducer that associates a determiner (optionally preceded by *à* or *de*) with a core noun phrase composed of a noun preceded by optional adjectives, in turn preceded by optional adverbs. Thus, sequences recognised by the grammar are retained by the parser only if they are (immediately or not) followed by a word that can be a noun. These experimental conditions are fair, since the test transducer corresponds to the conditions of use of the grammar.

[12] The tagging involves a linearization: when several recognized sequences overlap, or when a sequence is recognized by the grammar in several ways, for instance as a *Det* and as a *deDet*, the system arbitrarily chooses and tags only one of the sequences. A sequence may be recognized by the grammar but not tagged. This difference between the recognition capacity of the grammar and the result of the parsing, and the corresponding difference in recall, is an artefact of the method of evaluation, and not of the grammar.

These results show that the grammar is able to detect determiners with some accuracy, even on text which is not tagged for parts of speech[13]. Among the cases of wrong detection of determiners, 40% would be ruled out if the grammar checked agreement in number with the head noun and the uncountable feature of the head noun. This gives us two directions of further development of the grammar. Anyway, precision is less relevant than recall in the assessment of a component of a syntactic grammar. Among the cases where determiners were not recognised by the parser, the analysis of errors shows that many do not stem from the grammar but from other elements of the experimentation[14]. The 20% recall for the *deDet* graph stems mainly from the artefact mentioned in section 4.2. The grammar cannot discriminate whether a determiner is preceded by the preposition *de* or not. This is not a surprise, since the surface form *de* can be analysed either as a preposition, or as a determiner, or as a combination of a preposition and a determiner, and the choice depends on syntactic context.

## 6. Conclusion

We evaluated the quality of a grammar of French determiners by comparison with an independently annotated corpus. The application of the grammar gives deeper syntactic information than chunking or information available in treebanks: in particular, it contributes to a semantically more consistent detection of heads of noun phrases. The grammar achieves 86% precision and 92% recall, which is better than state of the art. The analysis of errors showed directions for improvement of both figures. These facts suggest that the local grammar is worth using as a component of a deep syntactic grammar of French.

---

[13] We also evaluated Silberztein's grammar in the same conditions as ours, which are not exactly the conditions for which it was constructed. We obtained 66% precision and 64% recall.

[14] 36% are explained by errors of lexical analysis (nouns not found in the lexicon or characters wrongly recognised as sentence boundaries), and another 12% by the fact that the test transducer did not allow quotation marks inside noun phrases. By solving such problems, one could probably obtain a recall of about 96% without updating the local grammar.